\documentclass{article}
\usepackage{ijcai24}

\usepackage{times}
\usepackage{soul}
\usepackage{url}
\usepackage[hidelinks]{hyperref}
\usepackage[utf8]{inputenc}
\usepackage[small]{caption}
\usepackage{graphicx}
\usepackage{amsmath}
\usepackage{amsthm}
\usepackage{booktabs}
\usepackage{algorithm}
\usepackage{algorithmic}
\usepackage[switch]{lineno}
\usepackage{multirow}
\usepackage{diagbox}

\title{PACE: A Pragmatic Agent for Enhancing Communication Efficiency Using Large Language Models
}

\author{
Jiaxuan Li$^{1,3,*}$\and
Minxi Yang$^{1,3,*}$\and
Dahua Gao$^{1,3}$\and
Wenlong Xu$^1$\and
Guangming Shi$^{1,2}$
\affiliations
$^1$the School of Artificial Intelligence, Xidian University, Xi'an, Shaanxi 710126 China\\
$^2$Peng Cheng Laboratory, Shenzhen, Guangdong 518055 China\\
$^3$Pazhou Lab (Huangpu), Guangzhou, Guangdong 510555 China\\
$^*$Equal contribution\\
\emails
jiaxuan\_li@stu.xidian.edu.cn,
mxyang@stu.xidian.edu.cn,
dhgao@xidian.edu.cn,
xuwl@stu.xidian.edu.cn,
gmshi@xidian.edu.cn
}

\begin{document}

\maketitle

\begin{abstract}
Current communication technologies face limitations in terms of theoretical capacity, spectrum availability, and power resources. Pragmatic communication, leveraging terminal intelligence for selective data transmission, offers resource conservation. Existing research lacks universal intention resolution tools, limiting applicability to specific tasks. This paper proposes an image pragmatic communication framework based on a Pragmatic Agent for Communication Efficiency (PACE) using Large Language Models (LLM). In this framework, PACE sequentially performs semantic perception, intention resolution, and intention-oriented coding. To ensure the effective utilization of LLM in communication, a knowledge base is designed to supplement the necessary knowledge, dedicated prompts are introduced to facilitate understanding of pragmatic communication scenarios and task requirements, and a chain of thought is designed to assist in making reasonable trade-offs between transmission efficiency and cost. For experimental validation, this paper constructs an image pragmatic communication dataset along with corresponding evaluation standards. Simulation results indicate that the proposed method outperforms traditional and non-LLM-based pragmatic communication in terms of transmission efficiency.
\end{abstract}

\begin{figure}[ht]

\centerline{\includegraphics[width=0.95\linewidth]{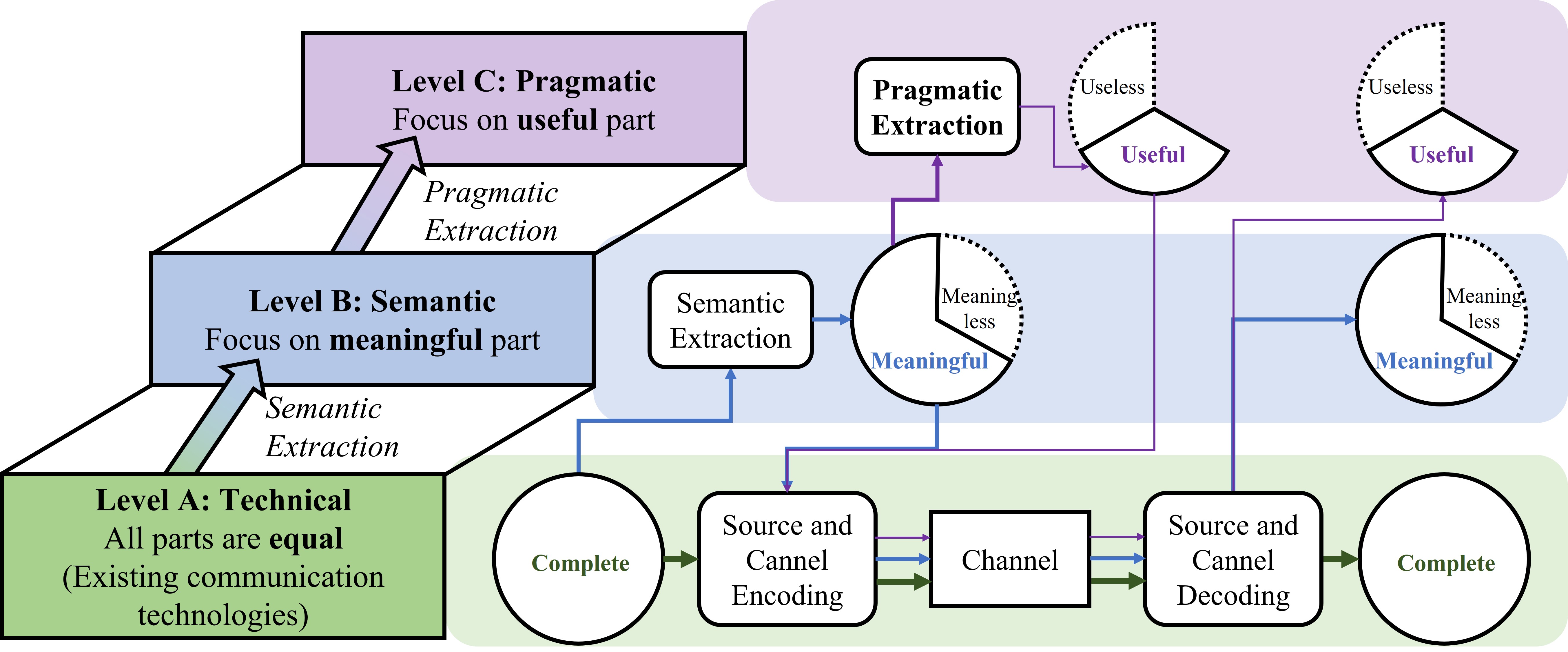}}
\caption{Weaver and Shannon categorized communication into three levels: technical, semantic, and efficiency (or pragmatic). The technical level focuses solely on symbol transmission, the semantic level emphasizes the transmission of the meaning represented by symbols, and the pragmatic level concerns the transmission of meaning in line with intentions. Semantic communication involves extracting meaning from symbols, removing meaningless parts to reduce transmission volume. Pragmatic communication, based on requirements, filters the semantics, performing pragmatic extraction to eliminate unnecessary portions and further reduce transmission volume. The thickness of the arrows in the diagram illustrates the difference in transmission volume.}
\label{fig:motivation}
\end{figure}

\section{Introduction}
Over the past decades, wireless communication technology has evolved from 1G, supporting only voice calls with a speed of 2 Kbps, to 5G with speeds exceeding 1 Gbps, enabling high-definition video transmission \cite{vora2015evolution}. However, this development model has faced constraints related to spectrum resources, transmission power, and the theoretical limits of coding, making it challenging to meet the demands of emerging technologies like extended reality, which require ultra-high-capacity communication \cite{chowdhury20206g}. 

In reality, not all data needs to be transmitted from the perspective of meeting user requirements. Weaver and Shannon categorized communication into technical, semantic, and efficiency (or pragmatic) levels \cite{weaver1953recent}, as illustrated in Figure \ref{fig:motivation}. The technical level is solely concerned with symbol transmission, the semantic level focuses on the transmission of the meaning represented by symbols, and the pragmatic level concentrates on the transmission of meaningful semantics in line with intentions. Semantic communication involves extracting meaning from symbols, eliminating meaningless parts to enhance communication efficiency \cite{shi2021from}. Pragmatic communication, driven by requirements, filters semantics through pragmatic extraction, discarding unnecessary components to further boost communication efficiency \cite{tung2021effective}. Pragmatic communication has emerged as a promising direction for the development of 6G, emphasizing the transmission of meaningful information aligned with user intention \cite{strinati20216g,rong20216g}. 

The key to successful pragmatic communication lies in the ability to accurately understand requirements and infer corresponding pragmatic extraction methods based on those needs. Currently, most pragmatic communication research relies on a specific downstream task (referred to as goal/task-oriented communication)\cite{gunduz2022beyond,strinati20216g}. In this approach, architecture is designed and parameters are optimized for semantic/pragmatic extraction and symbol encoding/decoding modules with task metrics as objectives. Due to the necessity of designing architectures and optimizing parameters for specific tasks, this approach exhibits limited task transferability and is challenging to apply in general scenarios. To overcome the constraints of specific tasks, a method is required for understanding and inferring general requirements. 

Recently, rapidly advancing Large Language Models (LLMs) \cite{brown2020language} have demonstrated remarkable comprehension and reasoning capabilities, giving rise to practical applications based on LLM-based agents \cite{xi2023rise}. LLMs are designed for text-based dialogues, and to utilize them for pragmatic communication, there is a need to describe task scenarios \cite{wang2023voyager,huang2023robust}, supplement necessary knowledge \cite{cui2023chatlaw,shi2023mededit}, and design Chain of Thought (CoT) \cite{kojima2022large,wang2022self} for them. 

This paper focuses on Image Pragmatic Communication (IPC) and proposes a parameter-free framework based on the LLM-Agent. The framework, named PACE (Pragmatic Agent for Communication Efficiency), treats the LLM-Agent as a communication node and revolves around the LLM to accomplish semantic perception, intention resolution, and encoding/decoding strategy planning. To ensure the LLM can effectively contribute to communication, the paper designs prompts to facilitate its understanding of pragmatic communication scenarios and requirements. Additionally, a knowledge base is devised to supplement the encoding/decoding knowledge, and a CoT is designed to assist in making reasonable trade-offs between transmission efficiency and cost. For experimental validation, the paper constructs an IPC dataset with corresponding evaluation criteria. Comparative experiments are conducted on transmission efficiency against traditional communication methods and pragmatic communication methods based on non-LLM approaches.

The contributions of this paper are summarized as follows: 
\begin{itemize}
    \item Proposing a training-free pragmatic communication framework based on the LLM-Agent for intention resolution and encoding/decoding strategy planning in image transmission, preliminary exploring the feasibility of general pragmatic communication.
    \item Designing specialized prompts, a knowledge base, and a CoT for the LLM-Agent to meet the requirements of IPC, allowing the LLM's general understanding and reasoning capabilities to be seamlessly transferred to pragmatic communication tasks without training.
    \item Establishing an IPC dataset with corresponding evaluation criteria, providing a benchmark for future research in IPC.
\end{itemize}

\begin{figure*}[ht]

\centerline{\includegraphics[width=0.95\linewidth]{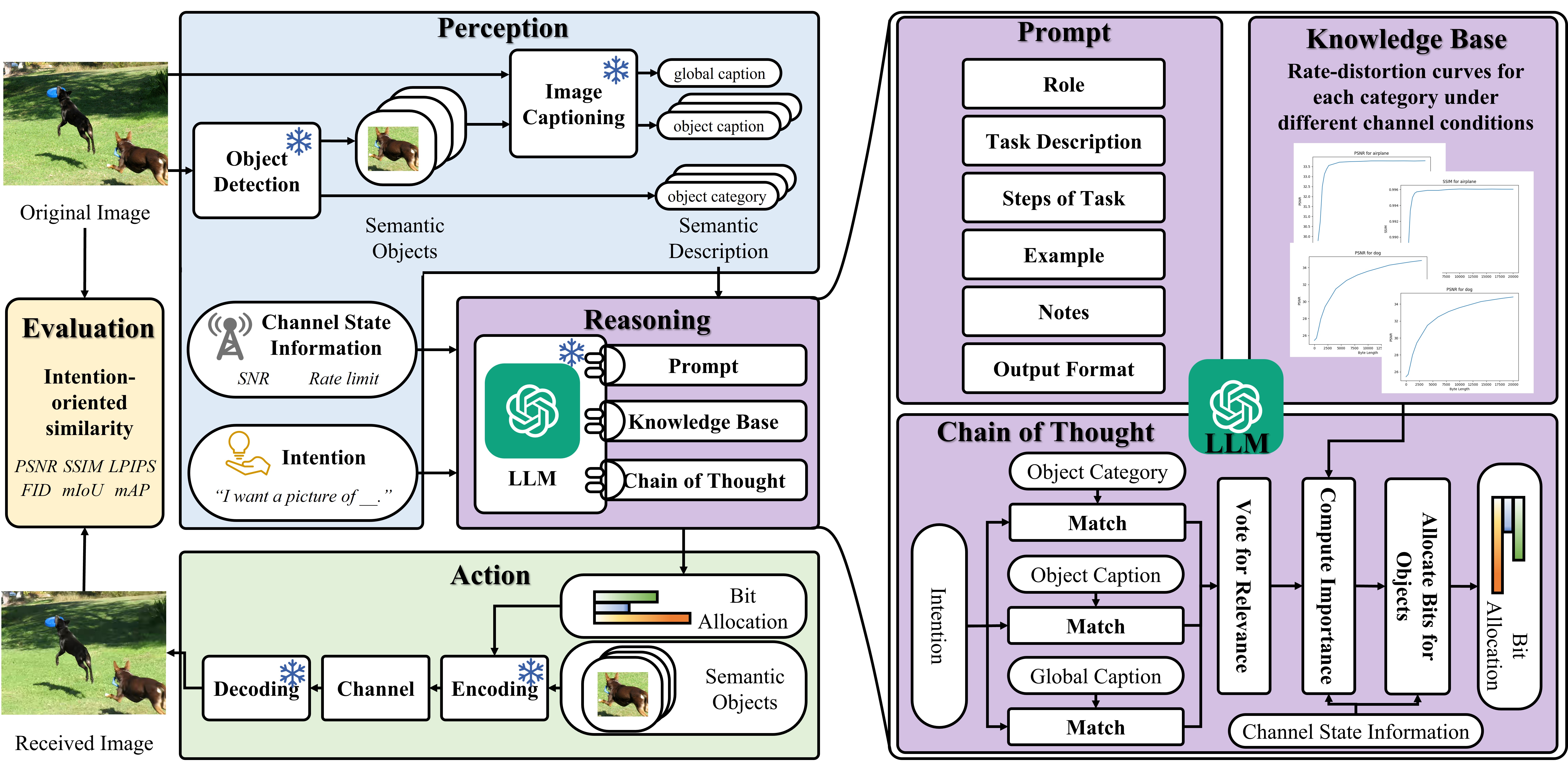}}
\caption{The proposed IPC framework based on PACE follows a process that includes image semantic perception, intention comprehension and reasoning, intent-adaptive image encoding and decoding, and intent-oriented evaluation. Modules marked with a snowflake indicate that no training is required for the entire process. }
\label{fig:overview}
\end{figure*}

 \section{Related Work}
\subsection{Pragmatic Communication}
The current research in the communication field at the pragmatic level is predominantly focused on situations where downstream tasks are well-defined, known as task/goal-oriented communication \cite{gunduz2022beyond}. The typical workflow involves optimizing semantic/pragmatic extraction, source-channel coding, and other components based on downstream task metrics. There are two main categories based on the representation of task objectives. 

One category treats downstream task evaluation standards as the objective function. It views the sender and receiver as encoder and decoder networks of an autoencoder, respectively. The optimization involves end-to-end joint optimization using differentiable channel models like Additive Gaussian White Noise (AWGN) channel. These methods require downstream task evaluation standards to be explicit and differentiable functions, making them more suitable for pragmatic communication oriented towards classic AI tasks such as classification \cite{kang2022task,yang2024superimposed}, matching \cite{jankowski2020deep}, detection \cite{yang2021semantic}, and question-answering \cite{xie2021task}. 

The other category represents task objectives as environments in reinforcement learning. Senders and receivers are implemented as agents, and reinforcement learning is employed to optimize these agents \cite{tung2021effective,tung2021context}. This approach does not require explicit differentiable channel models and evaluation functions, making it more suitable for real-world scenarios. However, it comes with the challenge of extended training times in complex real environments, as agents need to learn based on environmental feedback. 

The proposed method in this paper diverges from existing task-driven communication approaches. It leverages natural language to describe intentions and utilizes LLM for intention understanding and reasoning without the need for training. This preliminary exploration aims to assess the feasibility of general pragmatic communication. 

\subsection{LLM-Agent}
Recently, there has been rapid development in LLM, with ChatGPT as a representative example \cite{brown2020language,touvron2023llama,zeng2022glm}. Thanks to the outstanding comprehension and reasoning abilities of LLM \cite{creswell2022faithful}, along with its excellent transferability \cite{kojima2022large}, agents based on LLM, can perceive the environment, analyze reasoning, and make decisions in open-ended problems like video games without the need for specific training \cite{huang2023robust,wang2023voyager,park2023generative}. Some even argue that LLMs exhibit sparks of General Artificial Intelligence (AGI)  \cite{xi2023rise}.

LLMs are designed for text tasks, and when applying them to specific task scenarios, it often requires providing contextual information related to the task scenario in textual form. Contextual information can be achieved through the combined use of these three methods:
(1) Prompts: Describing the task scenario as fully as possible in prompts, including background, task definition, task objectives, input-output formats, and examples \cite{shen2023hugginggpt,white2023prompt}. Due to the limitations on the number of tokens in LLMs, prompts have limited capacity for information.
(2) Access to External Knowledge: Allowing LLMs to access an external knowledge base through queries \cite{cui2023chatlaw,shi2023mededit}. The external knowledge base can contain comprehensive information, and LLMs only need to retrieve the necessary portion through queries.
(3) Planning CoTs: Designing CoTs that adapt to the scenario for LLMs, breaking down a complex problem into simpler questions \cite{kojima2022large,wang2022self}. 

Inspired by these ideas, this paper designs specialized prompts, a knowledge base, and CoTs for the LLM-Agent in the context of IPC. This approach facilitates the transfer of LLM's general understanding and reasoning abilities to the task of pragmatic communication without the need for specific training. 

\section{Proposed Method}
This paper proposes an IPC framework based on PACE, as illustrated in Figure \ref{fig:overview}. The process involves four stages: image semantic perception, intention understanding and reasoning, intent-adaptive image encoding/decoding, and intent-oriented evaluation. Among them, intention understanding and reasoning based on LLM determine whether the subsequent encoding/decoding information meets the intention and communication conditions, making it crucial. To seamlessly transfer the LLM's general understanding and reasoning capabilities to pragmatic communication tasks without training, the paper designs specialized prompts, a knowledge base, and a CoT for PACE to meet the requirements of IPC. 
This section will first introduce the overall process of IPC based on PACE, followed by a detailed explanation of PACE's specialized prompts, knowledge base, and CoT. 

 \subsection{IPC Based on PACE}
As shown in Figure \ref{fig:motivation}, to selectively transmit useful data, pragmatic communication needs to undergo semantic extraction, pragmatic extraction, and encoding/decoding of data. This aligns perfectly with the agent's perception, reasoning, and action. In this paper, the information gathering phase, including semantic extraction of data, is considered as the perception module. The reasoning module is driven by LLM, handling information processing. The encoding, transmission, and decoding of data are treated as the action module. Intention-oriented similarity assessments are employed to evaluate the effectiveness of IPC, as illustrated on the left side of Figure \ref{fig:overview}. 

\subsubsection{Perception}
The perception module of PACE is responsible for collecting information, with a focus on crucial details contained in the data to be transmitted. To extract text information from image data that is understandable by LLM, PACE employs both object detection \cite{carion2020end} and image captioning \cite{li2022blip}. 
Given that natural language used in human communication is predominantly noun-centric, with nouns corresponding to objects in images, we utilize object detection to extract semantic information. Moreover, the rectangular image blocks obtained from object detection are more convenient for subsequent image encoding/decoding. 
The category names obtained from object detection serve as part of the information provided to LLM. Additionally, detailed descriptions of the detected objects, obtained through image captioning applied to the object regions, are also provided to LLM. Furthermore, the overall textual description of the entire image is also fed to LLM. This transformation of the image into text, understandable by LLM, encompasses diverse information ranging from global to local and from coarse to detailed. 

In communication, besides the data to be transmitted, there is also Channel State Information (CSI), such as the Signal-to-Noise Ratio (SNR) and rate limits. For pragmatic communication, it is essential to describe the requirements of the task goals. Leveraging LLM's text comprehension capabilities, PACE utilizes natural language to articulate intention. Compared to task-driven communication employing specific objective functions or environmental feedback, natural language intention is more flexible, simpler, more scalable, and closer to human habits. 

\subsubsection{Reasoning}
Communication is not merely about pursuing transmission efficiency but requires a balance between the effectiveness and costs of transmission. Under a given cost, maximizing the transmission effect by allocating limited resources to more critical parts, i.e., the resource allocation problem, is one of the key challenges in communication. 

The information provided to the reasoning module by the perception module includes the semantic description of the data, CSI, and intention. Here, the semantic description represents data features, CSI contains transmission conditions and cost constraints, and intention represents the purpose of transmission. Therefore, LLM needs to determine the importance of different semantics based on intention and semantic description, guided by CSI, to allocate resources. The outcome of resource allocation for an image includes the bounding boxes of all objects along with their corresponding bit allocations, as well as the bit allocation for the background. 

To accomplish this process, this paper designs specialized prompts for PACE to clarify the scene and task requirements, constructs a knowledge base to supplement the required communication knowledge, and develops a CoT to simplify the problem. These aspects will be detailed in subsection \ref{subsec:plugins}. 

\subsubsection{Action}
The action module performs encoding, transmission, and decoding based on the allocated resources for each object. Since PACE does not require joint optimization of the encoder and other modules, the implementation of the encoder and channel is not restricted. In this work, classic encoding methods are employed, with Better Portable Graphics (BPG) used for source coding, Low-Density Parity Check
codes (LDPC) \cite{richardson2018design} for channel coding, and AWGN for the channel. 

During encoding, the image blocks within each object bounding box are sequentially subjected to source and channel coding based on the allocated bit lengths. Subsequently, the regions of all objects are filled with pure white color to represent the background, and source and channel coding are applied based on the bit lengths of the background. 

During transmission, the background is transmitted first, followed by the sequential transmission of all objects based on the order of their bounding box center points. 

During decoding, the background is decoded first, and then, based on the received order of objects, each decoded object is filled into the corresponding position in the background. 

\subsubsection{Evaluation} \label{sec:eval}
This paper introduces an intention-oriented similarity metric as the evaluation criterion for IPC, building upon classical image evaluation standards. Two categories of classical evaluation standards are employed. 

The first category includes image similarity metrics such as Peak Signal-to-Noise Ratio (PSNR), Structural SIMilarity (SSIM) \cite{wang2004image},  Learned Perceptual Image Patch Similarity (LPIPS) \cite{zhang2018unreasonable}, and Fréchet inception distance (FID), which measure the detailed quality of images. 
PSNR is a metric used to evaluate the quality of image or video reconstruction.  
SSIM comparing their similarities in terms of luminance, contrast, and structure. These two metrics primarily assess the quality of reconstructed images at the pixel level. The higher these two indicators, the better the quality of the reconstructed image. 
FID calculates the similarity between generated and real images by extracting feature vectors from each image and computing their Fréchet distance. 
LPIPS evaluates the similarity between images by comparing their perceived differences with human perception results. Both metrics assess the quality of reconstructed images at the perceptual level, with lower values indicating higher reconstruction quality. 

The second category consists of evaluation standards used in image understanding tasks, including mean-Average-Precision (mAP) for object detection \cite{carion2020end} and mean-Intersection-over-Union (mIoU) for segmentation \cite{cheng2021per}, which assess the semantic accuracy of images. 
mAP is obtained by performing object detection on the reconstructed images.
mIoU is obtained by conducting semantic segmentation on the reconstructed images. These two metrics assess the performance of our method at the downstream task level. Higher values of mAP and mIoU indicate better quality of the reconstructed images.

For the evaluation based on the first category, the similarity is computed only for the original image and the received image at the locations corresponding to the objects that satisfy the intention. The evaluation based on the second category involves using the same object detection/semantic segmentation model for both the original and received images. The parts in the predicted result of the original image that satisfy the intention serve as the ground truth, and the predicted result of the received image is then evaluated accordingly.

\begin{table*}[h]
\centering
\caption{Comparison results between PACE and methods based on text similarity and CLIP. The labels "Global," "Intention Matching Level 1," "Intention Matching Level 2," and "Intention Matching Level 3" respectively indicate the metrics for the entire image, regions with matching levels 1, 2, and 3. For the intention-oriented metrics, higher values are desirable when the intention is satisfied (matching level 3), while lower values are preferred when the intention is not satisfied (matching levels 1 and 2).}
\label{tab1}
\resizebox{2.0\columnwidth}{!}{
\renewcommand\arraystretch{2}
\Huge
\begin{tabular}{cccccccccccccccccccccc}
\hline
\multicolumn{1}{c}{\multirow{2}{*}{\textbf{Dataset}}} & \multicolumn{1}{c}{\multirow{2}{*}{\textbf{Method}}} & \multicolumn{2}{c}{\textbf{Global}} & \multicolumn{6}{c}{\textbf{Intention Matching Level 3}}  & \multicolumn{6}{c}{\textbf{Intention Matching Level 2}}& \multicolumn{6}{c}{\textbf{Intention Matching Level 1}}\\

\cmidrule(r){3-4}
\cmidrule(r){5-10}
\cmidrule(r){11-16}
\cmidrule(r){17-22}

\multicolumn{1}{c}{} & \multicolumn{1}{c}{} & \textbf{PSNR}$\uparrow$ & \textbf{SSIM}$\uparrow$ & \textbf{PSNR}$\uparrow$& \textbf{SSIM}$\uparrow$ & \textbf{FID}$\downarrow$ & \textbf{LPIPS}$\downarrow$ & \textbf{mAP}$\uparrow$ & \textbf{mIoU}$\uparrow$ & \textbf{PSNR}$\downarrow$ & \textbf{SSIM}$\downarrow$ & \textbf{FID}$\uparrow$ & \textbf{LPIPS}$\uparrow$ & \textbf{mAP}$\downarrow$ & \textbf{mIoU}$\downarrow$ & \textbf{PSNR}$\downarrow$ & \textbf{SSIM}$\downarrow$ & \textbf{FID}$\uparrow$ & \textbf{LPIPS}$\uparrow$ & \textbf{mAP}$\downarrow$ & \textbf{mIoU}$\downarrow$ \\ 
\hline
\multirow{4}{*}{COCO} 
                      & CLIP & \textbf{29.71}  & \textbf{0.83}  & 33.45  & 0.90  & 1.31  & 0.10  & \textbf{0.50}
& 0.52  & 29.73  & 0.86  & 1.61  & 0.13  & 0.39  & \textbf{0.44}  & 34.18  & 0.90  & 1.49  & 0.10  & 0.43  & 0.59  \\
                      & Text-sim & 29.61  & \textbf{0.83}  & 33.62  & 0.89  & 1.34  & 0.11  & 0.47
& 0.52  & 29.30  & 0.85  & \textbf{1.65}  & 0.14  & 0.47  & 0.46  & 32.30  & 0.88  & 1.62  & 0.12  & 0.43  & 0.53  \\
             & PACE (Ours) & 29.66  & \textbf{0.83}  & \textbf{34.71}  & \textbf{0.91}  & \textbf{1.04}  & \textbf{0.09}  & 0.49
& \textbf{0.57}  & \textbf{28.02}  & \textbf{0.83}  & 1.49  & \textbf{0.15}  & \textbf{0.35}  & 0.46  & \textbf{29.17}  & \textbf{0.84}  & \textbf{1.74}  & \textbf{0.16}  & \textbf{0.40}  & \textbf{0.46}  \\                    
\hline

\multirow{4}{*}{Flicker8K} 
                      & CLIP & \textbf{28.92}  & \textbf{0.83}  & 31.69  & \textbf{0.90}  & 1.47  & 0.10  & 0.36
& 0.37  & 29.61  & 0.86  & 1.64  & 0.13  & 0.36  & 0.31  & 31.76  & 0.91  & 1.38  & 0.10  & 0.41  & 0.46  \\ 
                      & Text-sim & 28.55  & 0.82  & 31.42  & 0.88  & 1.67  & 0.11  & 0.41
& 0.37  & 29.70  & 0.85  & \textbf{1.83}  & 0.13  & 0.30  & \textbf{0.29}  & 29.69  & 0.86  & 1.65  & 0.14  & \textbf{0.36}  & 0.43  \\ 
             & PACE (Ours) & 28.85  & \textbf{0.83}  & \textbf{32.56}  & \textbf{0.90}  & \textbf{1.22}  & \textbf{0.09}  & \textbf{0.47}
& \textbf{0.40}  & \textbf{28.28}  & \textbf{0.83}  & 1.46  & \textbf{0.15}  & \textbf{0.29}  & 0.31  & \textbf{28.12}  & \textbf{0.85}  & \textbf{1.71}  & \textbf{0.15}  & 0.37  & \textbf{0.37}  \\ 
\hline

\multirow{4}{*}{Flicker30K} 
                      & CLIP & 28.32  & \textbf{0.83}  & 32.63  & 0.91  & 1.19  & 0.09  & 0.52
& 0.47  & 30.45  & 0.88  & \textbf{1.28}  & 0.11  & 0.26  & 0.41  & 32.04  & 0.90  & 1.28  & 0.10  & 0.36  & 0.51  \\
                      & Text-sim & 28.05  & \textbf{0.83}  & 32.52  & 0.90  & 1.22  & 0.10  & 0.50
& 0.48  & \textbf{29.37}  & \textbf{0.86}  & 1.35  & \textbf{0.13}  & 0.24  & 0.38  & 31.10  & 0.87  & 1.42  & 0.12  & \textbf{0.33}  & 0.47  \\ 
             & PACE (Ours) & \textbf{29.66}  & \textbf{0.83}  & \textbf{33.70}  & \textbf{0.92}  & \textbf{0.97}  & \textbf{0.08}  & \textbf{0.57}
& \textbf{0.52}  & 30.08  & 0.87  & 1.13  & \textbf{0.13}  & \textbf{0.16}  & \textbf{0.37}  & \textbf{28.03}  & \textbf{0.83}  & \textbf{1.59}  & \textbf{0.16}  & 0.37  & \textbf{0.35}  \\ 
\hline

\end{tabular}
}
\end{table*}

\begin{table*}[h]
\centering
\caption{Ablation study results for PACE. "W/O voting" and "W/O knowledge base" represent the scenarios where the voting-based matching calculation in CoT and the knowledge base are excluded, respectively. Higher intention-oriented metrics are desirable when the intention is satisfied (matching level 3), while lower values are preferred when the intention is not satisfied (matching levels 1 and 2). }
\label{tab2}
\resizebox{2\columnwidth}{!}{
\renewcommand\arraystretch{2}
\small
\begin{tabular}{cccccccccccccccc}
\hline
\multicolumn{1}{c}{\multirow{2}{*}{\textbf{Ablation}}} & \multicolumn{2}{c}{\textbf{Global}} & \multicolumn{6}{c}{\textbf{Intention Matching Level 3}}  & \multicolumn{6}{c}{\textbf{Intention Matching Level 1\&2}}\\
\cmidrule(r){2-3}
\cmidrule(r){4-9}
\cmidrule(r){10-15}

\multicolumn{1}{c}{} & \multicolumn{1}{c}{\textbf{PSNR}$\uparrow$} & \multicolumn{1}{c}{\textbf{SSIM}$\uparrow$} & \multicolumn{1}{c}{\textbf{PSNR}$\uparrow$} & \multicolumn{1}{c}{\textbf{SSIM}$\uparrow$} & \multicolumn{1}{c}{\textbf{FID}$\downarrow$} & \multicolumn{1}{c}{\textbf{LPIPS}$\downarrow$} & \multicolumn{1}{c}{\textbf{mAP}$\uparrow$} & \multicolumn{1}{c}{\textbf{mIoU}$\uparrow$} & \multicolumn{1}{c}{\textbf{PSNR}$\downarrow$} & \multicolumn{1}{c}{\textbf{SSIM}$\downarrow$} & \multicolumn{1}{c}{\textbf{FID}$\uparrow$} & \multicolumn{1}{c}{\textbf{LPIPS}$\uparrow$} & \multicolumn{1}{c}{\textbf{mAP}$\downarrow$} & \multicolumn{1}{c}{\textbf{mIoU}$\downarrow$}\\ 
\hline 
Normal  & \textbf{29.66}  & \textbf{0.83}  & \textbf{34.71} & \textbf{0.91} & \textbf{1.04} & \textbf{0.09} & \textbf{0.49} & \textbf{0.57} & 
28.60 & 0.84 & 1.62 & 0.15 & \textbf{0.38} & 0.46\\ 

W/O voting & 29.63  & \textbf{0.83}  & 33.36  & 0.89  & 1.30  & 0.11  & 0.45 & 0.48  & 
\textbf{27.59}  & \textbf{0.79}  & \textbf{1.92}  & \textbf{0.19}  & 0.39  & \textbf{0.40} \\

W/O knowledge base & 29.61  & \textbf{0.83}  & 34.70  & \textbf{0.91}  & 1.07  & \textbf{0.09}  & 0.47 & 0.53  & 
29.64  & 0.85  & 1.56  & 0.14  & 0.39  & 0.46 \\                 
\hline

\end{tabular}
}
\end{table*}

\begin{figure*}[ht]

\centerline{\includegraphics[width=0.95\linewidth]{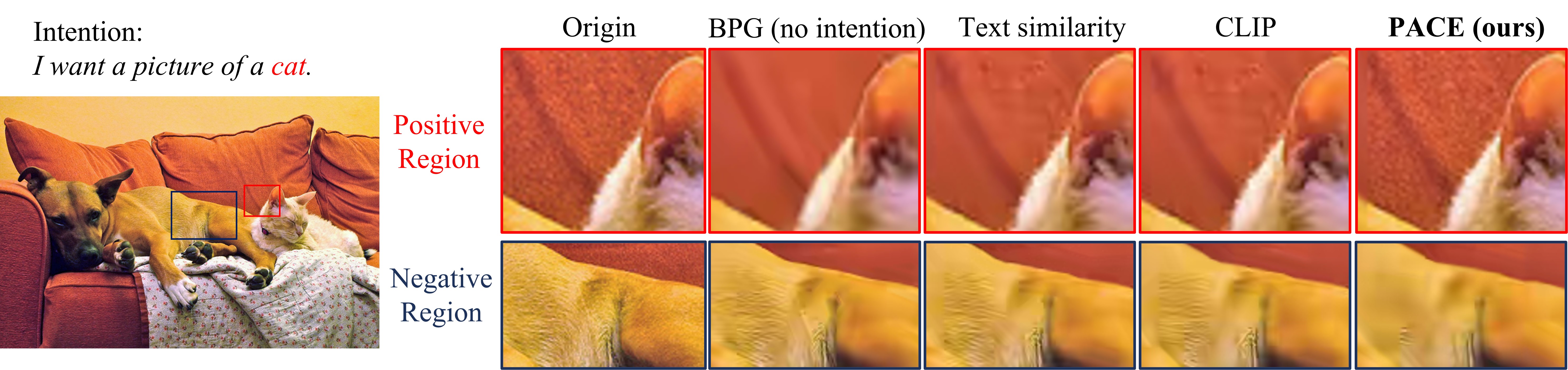}}
\caption{The comparison involves PACE, intention-agnostic, text-similarity-based, and CLIP-based methods. Regions within the red boxes correspond to satisfying the intention (positive regions), while the black regions represent not satisfying the intention (negative regions).}
\label{fig:sample}
\end{figure*}
\begin{figure*}[ht]

\centerline{\includegraphics[width=0.95\linewidth]{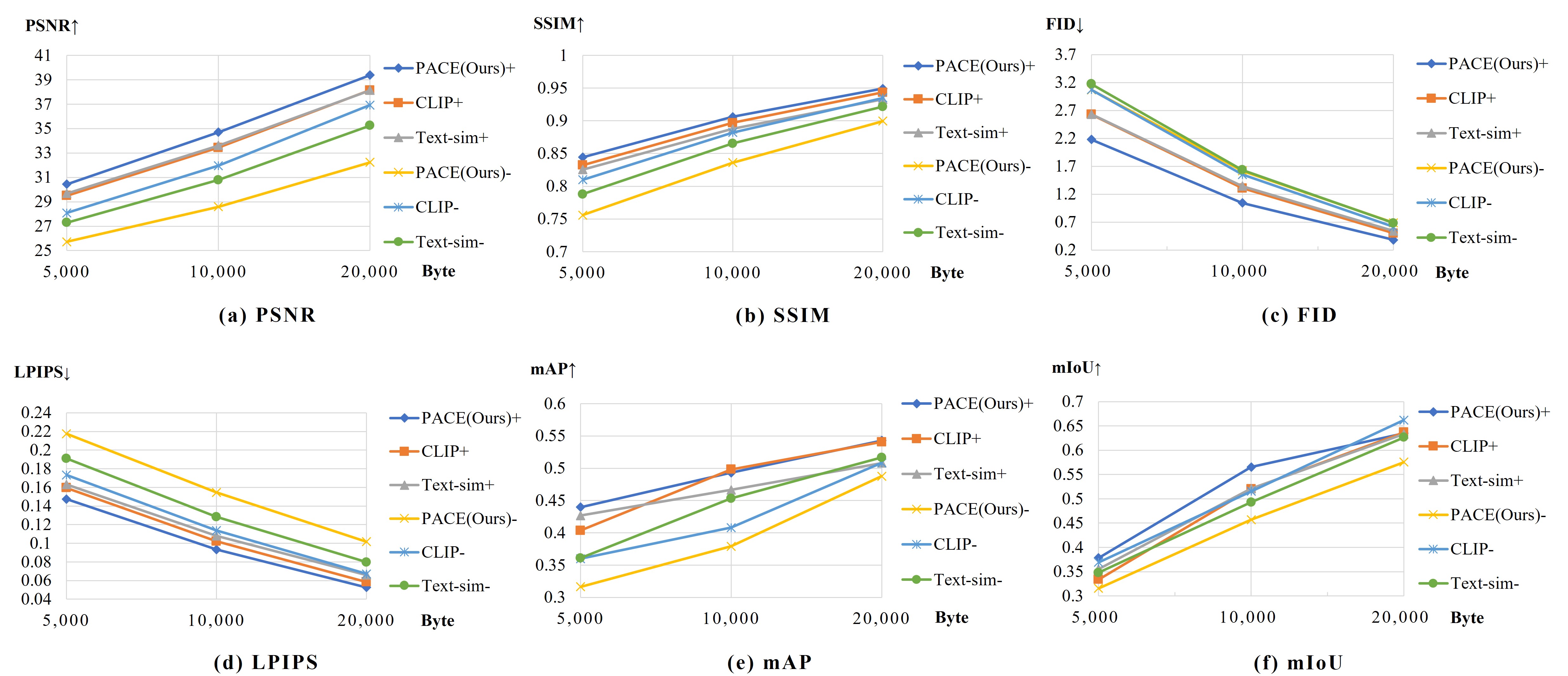}}
\caption{The line chart depicting the impact of bit length, where the plus and minus signs in the method names represent the metrics when matching (level 3) and not matching (level 1\&2) with the intention, respectively.}
\label{fig:effect_of_byte_length}
\end{figure*}

\subsection{Prompts, Knowledge Base and CoT}\label{subsec:plugins}
\subsubsection{Prompts}
The prompt framework employed by PACE is depicted in Figure \ref{fig:overview} and encompasses role definition, task description, task steps, examples, notes, and output format. Detailed prompts can be found in the appendix. 

\subsubsection{Knowledge Base}
The encoding difficulty varies for objects of different categories. For instance, compared to the surface of an airplane, a dog's fur has finer and more irregular textures. Under different channel conditions, PACE records the average image similarity (such as PSNR) corresponding to different bit lengths for different object categories. The resulting rate-distortion curves, as illustrated in Figure \ref{fig:overview}, serve as PACE's knowledge base. 
PACE adapts resource allocation for communication conditions and objects based on the desired similarity, channel conditions, and objects. It identifies the required bit lengths from the corresponding curves to achieve communication condition and object-adaptive resource allocation. 

\subsubsection{Chain of Thought}
To derive the resource allocation scheme for objects based on semantic description, CSI, and intention, PACE's CoT is depicted in Figure \ref{fig:overview}. 

\textbf{Calculate relevance.} The global caption, object category, and object caption are independently matched with the intention. The matching results are categorized into high, medium, and low, representing full, partial, and no match. Combine the results from the three parts to obtain the correlation between the object and intention. The correlation is calculated by assigning scores of 6, 4, and 2 to high, medium, and low matches for object category. Global caption and object caption results act as modifiers with scores of +1, 0, and -1 for high, medium, and low matches. The scores from the three matching results are summed to get a correlation value in the range of [0, 8].

\textbf{Calculate importance.} Use CSI and the knowledge base to calculate the importance of the object. For each correlation value, set the corresponding expected image similarity in the knowledge base. The higher the correlation, the higher the expected image similarity. Query the knowledge base to obtain the corresponding bit length. The object's bit length is the ratio of its bit length to the total bit length of all objects in the image, representing object importance. 

\textbf{Resource allocation.} Allocate bit lengths to each object based on the total bit length. Set the object's bit length as the sum of two parts. Considering that higher-resolution images require longer code lengths, use the ratio of the object area to the total image area and the average object importance as a resource allocation factor. Multiply the resource allocation factor by the total bit length to obtain the bit length for each object. 

\section{Experiments}
\subsection{Setup}
\label{4.1}
\subsubsection{Dataset}
We performed experiments on the validation set of the COCO2017 dataset \cite{lin2014microsoft}, as well as the test sets of the Flickr8k dataset \cite{hodosh2013framing} and the Flickr30k dataset \cite{plummer2015flickr30k}. To generate the intentions for these datasets, we initially employed the method proposed by \cite{carion2020end} to obtain object labels for each image in the dataset. In order to precisely determine the degree of alignment between the intent and the data description, we assigned a one-third probability for each label in an image to remain unchanged, indicating a matching level of 3. Similarly, there was a one-third probability for it to be replaced with a semantically similar word, resulting in a matching level of 2, and a one-third probability for a label to be deleted, reflecting a matching level of 1. Finally, we merged the processed labels with a prompt and concatenated them to form the intention for each image.

\subsubsection{Metric} 
The experiments utilized intention-oriented PSNR, SSIM, FID, LPIPS, mAP, and mIoU as evaluation metrics (refer to Section \ref{sec:eval}, Evaluation). For mAP and mIoU, DETR \cite{carion2020end} and MaskFormer \cite{cheng2021per} were employed for object detection and semantic segmentation on the received images. Ground truth used for calculating mAP and mIoU were selected from the original datasets' ground truth with matching level. In the subsequent presentation of experimental results, higher intention-oriented similarity metrics are desirable for matching level 3, while lower values are preferable for matching levels 1 and 2. 

\subsubsection{LLM}
We leveraged OpenAI's gpt-3.5-turbo-16k \cite{brown2020language} API for bit allocation. For LLM, we set temperature=0 and topk=1.

\subsubsection{Channel}
We conducted experiments on an AWGN channel with a SNR set at 20. For the modulation scheme, we used Quadrature Amplitude Modulation (QAM)-16. As for the channel encoding, we employed LDPC coding \cite{richardson2018design}, with parameters k=3072 and n=6144. where k represents the number of data bits per LDPC codeword and n represents the total number of bits in each LDPC codeword.

\subsubsection{Training}
Our method did not require training, the caption model and detection model used are both pretrained models with their weights frozen in PACE.

\subsubsection{Hardware and Software Platform}
All experiments were conducted on a desktop computer with an Intel I9 10920X@3.5GHz processor, 128GB of memory, and an NVIDIA RTX 3090 GPU. The operating system used was Ubuntu 18.04, and the CUDA version was 11.4. The target detection \cite{carion2020end} and image captioning \cite{li2022blip} in PACE utilized the code provided by the authors, while the implementation of encoding, decoding, and channel transmission was based on Sionna \cite{hoydis2022sionna}.

\subsection{Comparison with Other Methods}
Due to the lack of readily available methods suitable for pragmatic communication, we made diligent efforts to select a few representative algorithms as baselines. These methods were originally designed solely for calculating image-text similarity and had not been applied in the context of pragmatic communication before. We re-implemented them to ensure their functionality in pragmatic communication scenarios and compatibility with our experimental setup.

\textbf{BPG+LDPC}: We used the state-of-the-art image compression method BPG as the source coding technique. The encoding process does not involve any intention. 

\textbf{CLIP}: CLIP \cite{radford2021learning} is a method for learning the similarity between images and text, enabling direct computation of their similarity score. We calculate the similarity between the detected object bounding boxes and the intentions by using CLIP. The allocation of bits is determined based on the computed visual-textual similarity. 

\textbf{Text Similarity}: We use \cite{wang2020minilm} to map data description and intentions to a 384 dimensional dense vector space, then use cosine similarity to compute the degree of matching between images and intentions for bit allocation.

We compared the effectiveness of our approach with other methods under a channel width of 10,000, which represents that the byte length transmitted through the channel for each image does not exceed 10,000. In Table \ref{tab1}, we presented the results of our method and compared it with other approaches on three datasets. Due to the fact that the bit allocation-based approach assigns more bits to the target of intention, the number of bits allocated to the background is relatively low. However, when calculating the overall PSNR and SSIM of the entire image, the background has a significant influence. As a result, the bit allocation-based approach yields lower PSNR and SSIM values for the overall image compared to the BPG+LDPC method. The PSNR and SSIM values for image reconstruction using the BPG+LDPC method on the COCO2017 dataset were found to be 32.97 dB and 0.86, respectively. On the Flickr8k dataset, the PSNR and SSIM values were 32.23 dB and 0.87, respectively. Similarly, on the Flickr30k dataset, the PSNR and SSIM values were 31.85 dB and 0.87, respectively. Our method surpassed other methods in all indicators for targets with a match level 3. For targets with match levels 2 or 1, our method's indicators were lower than those of other methods, illustrating that our approach accurately understood the intention and selectively allocated bits based on that intention. This enabled more precise transmission of intention-relevant region while allowing relatively fuzzy transmission for region that were less important to the intention.

Figure \ref{fig:sample} presents the reconstructed images obtained using ours and other approaches. Under the given intention, ours achieves higher bits allocation for image regions that match the intended content, resulting in better preservation of details and textures after communication. Conversely, image regions that do not match the intention are assigned fewer bits, leading to a blurrier representation of texture details.

\subsection{Ablation Study}
In this section, we investigate the main components of PACE through ablation studies to observe their impact on performance. All experiments are conducted on the COCO2017 dataset \cite{lin2014microsoft}.

\subsubsection{CoT} 
We removed the proposed CoT and instead, for the same information and samples, had the LLM directly score the data descriptions and intentions on an score of 0-8. We then allocated the bits based on the output scores. In Table \ref{tab2}, W/O voting represents not using CoT. We observed that the absence of CoT results in a decrease in performance for all match levels of the objects. This is because LLM tends to assign a fixed score when scoring the degree of matching, thereby reducing the discriminability of the object from the background and other objects.

\subsubsection{Knowledge Base} 
In this experiment, we assigned a fixed bit length to each matching score for all targets. The bits length was no longer obtained by querying the knowledge base. In Table \ref{tab2}, W/O knowledge base represents the condition where the knowledge base was not used. Due to the varying number of bits required for different target transfers, using a fixed bit length failed to leverage inter-class differences, resulting in inefficient bit allocation.

\subsubsection{Byte Length} 
Figure \ref{fig:effect_of_byte_length} presents a comparison of the image reconstruction quality between ours and other approaches under different byte lengths. This comparison enables a better assessment of the effectiveness of our proposed bit allocation strategy in various channel conditions. Across all metrics, ours consistently allocates more bits to the region of interest compared to other methods, which display a lower degree of discrimination. It is worth noting that most metrics indicate that at lower byte lengths, our method achieves higher reconstruction quality for positive region. This observation highlights the ability of our method to allocate limited bits to the region that the receiver requires the most.

\section{Conclusion}
In this study, we introduced PACE, an untrained architecture utilizing LLM-Agent for IPC. PACE seamlessly integrates LLM capabilities for semantic perception, intention resolution, and intention-oriented coding. Prompts, knowledge base, and CoT are designed to facilitate LLM in harnessing its understanding and reasoning abilities in IPC. 
To carry out experiments for PACE, an IPC dataset and intent-oriented evaluation standards are proposed.
Experimental comparisons with traditional and non-LLM-based methods validate the effectiveness of PACE.  Based on these efforts, this study provides an initial exploration into the feasibility of universal pragmatic communication.

\vfill\pagebreak
\bibliographystyle{named}
\bibliography{main}
\clearpage

\appendix


\begin{figure*}[ht]
\setcounter{figure}{0}
\renewcommand{\thefigure}{A\arabic{figure}}
\centerline{\includegraphics[width=0.95\linewidth]{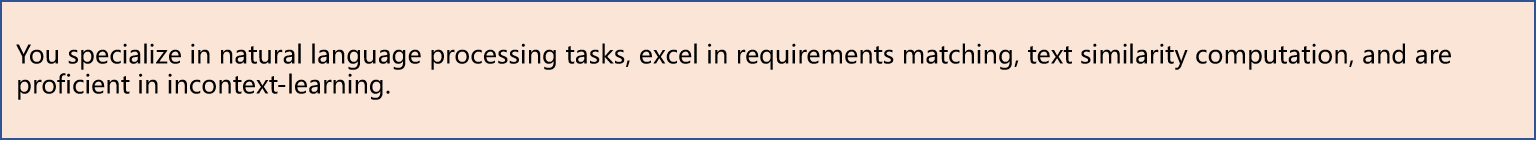}}
\caption{This prompt defines the role that LLM plays in this task.}
\label{fig:prompt}
\end{figure*}
\begin{figure*}[ht]
\renewcommand{\thefigure}{A\arabic{figure}}
\centerline{\includegraphics[width=0.95\linewidth]{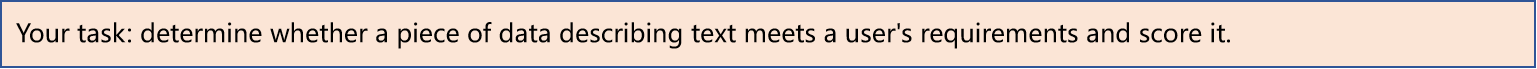}}
\caption{This prompt defines the tasks that LLM needs to complete.}
\label{fig:prompt}
\end{figure*}
\begin{figure*}[ht]
\renewcommand{\thefigure}{A\arabic{figure}}
\centerline{\includegraphics[width=0.95\linewidth]{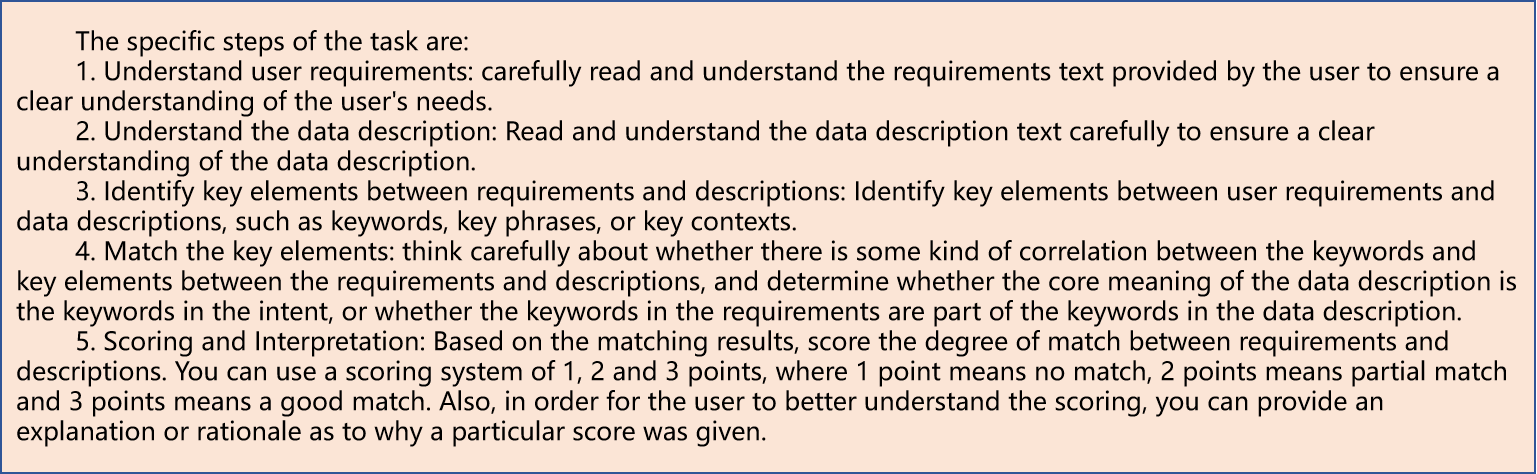}}
\caption{This prompt defines the specific steps required for LLM to complete the task.}
\label{fig:prompt}
\end{figure*}
\begin{figure*}[ht]
\renewcommand{\thefigure}{A\arabic{figure}}
\centerline{\includegraphics[width=0.95\linewidth]{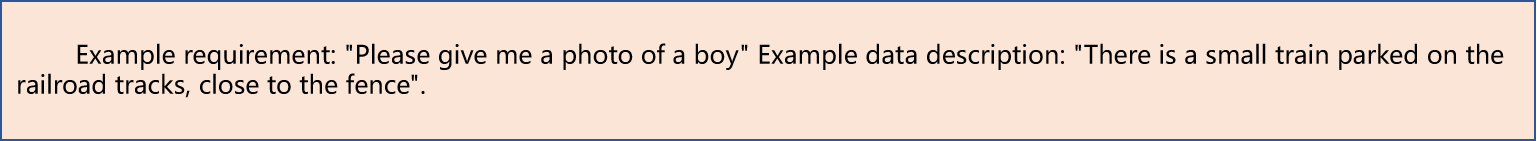}}
\caption{This prompt shows an example of an input.}
\label{fig:prompt}
\end{figure*}
\begin{figure*}[ht]
\renewcommand{\thefigure}{A\arabic{figure}}
\centerline{\includegraphics[width=0.95\linewidth]{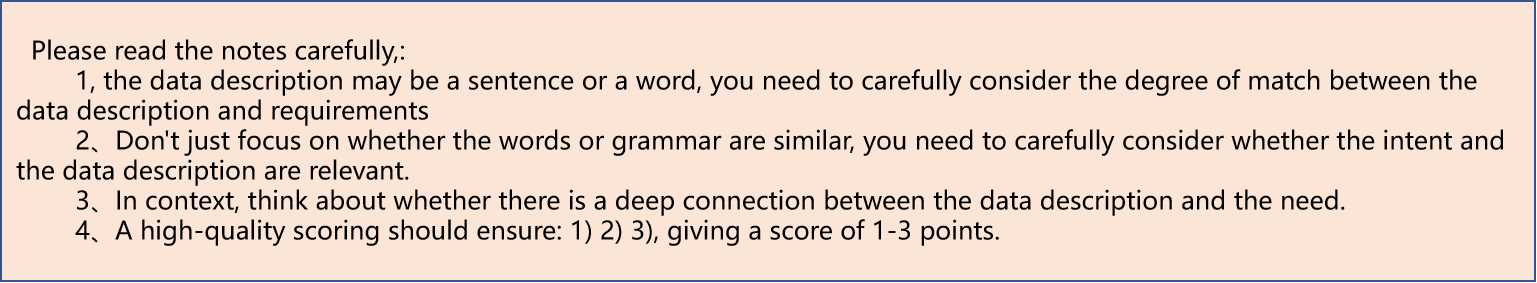}}
\caption{This prompt defines the precautions that LLM needs to pay attention to when completing tasks.}
\label{fig:prompt}
\end{figure*}
\begin{figure*}[ht]
\renewcommand{\thefigure}{A\arabic{figure}}
\centerline{\includegraphics[width=0.95\linewidth]{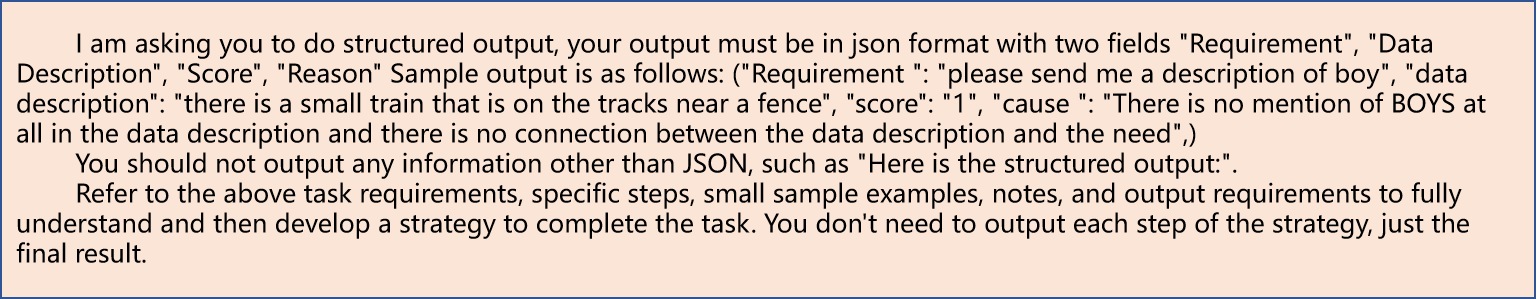}}
\caption{This prompt restricts the output format of LLM.}
\label{fig:prompt}
\end{figure*}

\begin{figure*}[ht]
\renewcommand{\thefigure}{A\arabic{figure}}
\centerline{\includegraphics[width=0.80\linewidth]{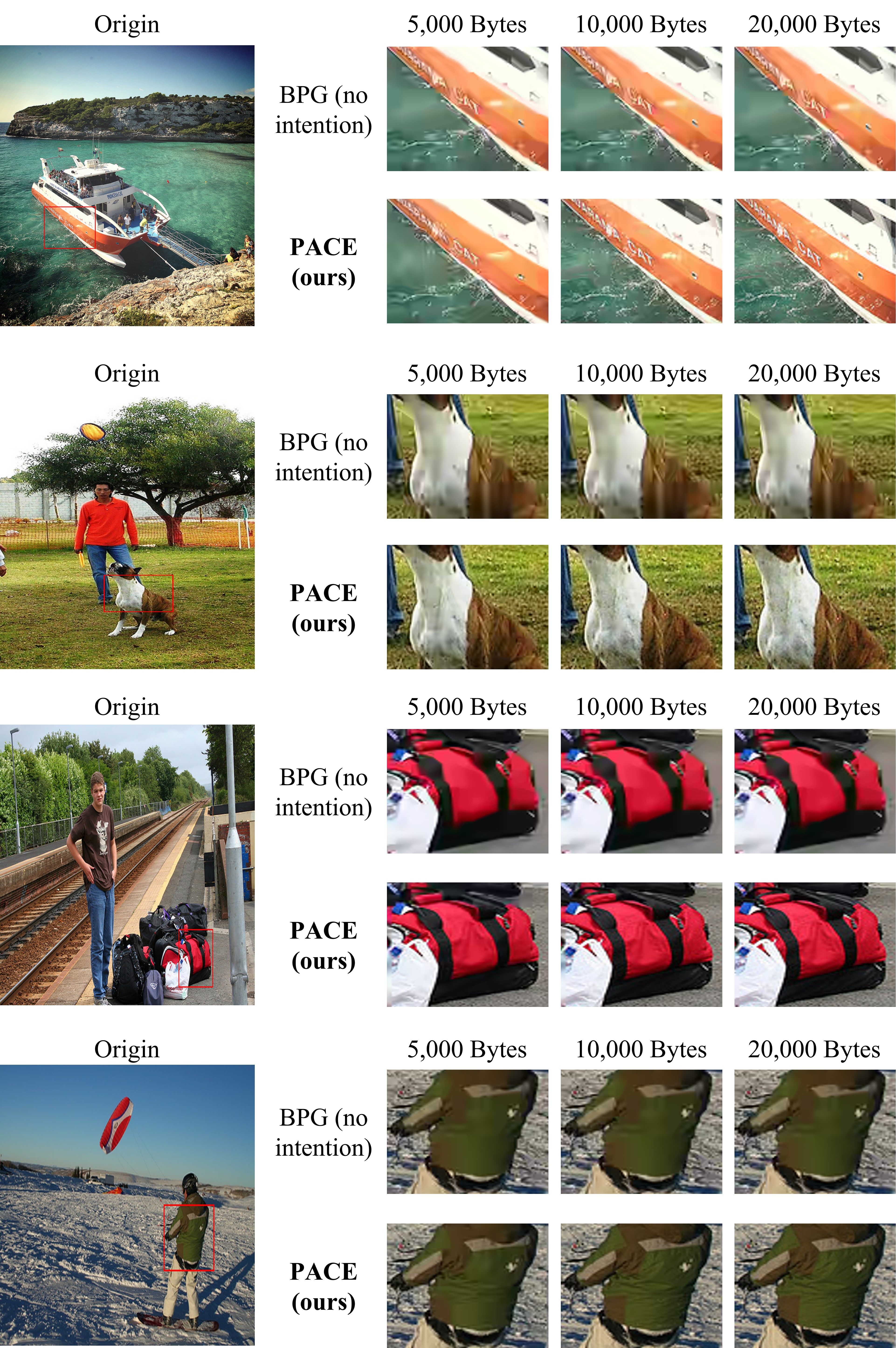}}
\caption{Samples at various bit lengths. Regions within the red boxes correspond to satisfying the intention.}
\label{fig:more_samples}
\end{figure*}
\end{document}